  \documentclass[final,5p,times]{elsarticle}

\usepackage{amssymb}
\usepackage{lipsum}
\usepackage{amsmath} 
\usepackage{todonotes}

\journal{}

\begin{document}

\begin{frontmatter}



\title{Shape-preserving Tooth Segmentation from CBCT Images Using Deep Learning with Semantic and Shape Awareness}


\author[one,two]{Zongrui Ji}
\author[three1]{Zhiming Cui}
\author[three]{Na Li}
\author[four]{Qianhan Zheng}
\author[one]{Miaojing Shi}
\author[five]{Ke Deng}
\author[six]{Jingyang Zhang}
\author[seven]{Chaoyuan Li}
\author[four]{Xuepeng Chen}
\author[one,two]{Yi Dong\corref{cor1}}
\cortext[cor1]{Corresponding author.}
\ead{yidong@tongji.edu.cn}
\author[one,two]{Lei Ma\corref{cor1}}
\ead{ma_lei@tongji.edu.cn}

\affiliation[one]{organization={Department of Control Science and Engineering, School of Electronics and Information Engineering, Tongji University},
            city={Shanghai},
            postcode={201804}, 
            country={China}}

\affiliation[two]{organization={Shanghai Institute of Intelligent Science and Technology, Tongji University},
            city={Shanghai},
            postcode={200092}, 
            country={China}}   
            
\affiliation[three1]{organization={School of Biomedical Engineering \& State Key Laboratory of Advanced Medical Materials and Devices, ShanghaiTech University},
            city={Shanghai},
            postcode={201210}, 
            country={China}}  

\affiliation[three]{organization={Stomatological Hospital of Henan Province, The First Affiliated Hospital of Zhengzhou University, School and Hospital of Stomatology of Zhengzhou University},
            city={Zhenzhou},
            postcode={450052}, 
            country={China}}

\affiliation[four]{organization={Stomatology Hospital, School of Stomatology, Zhejiang University School of Medicine, Clinical Research Center for Oral Diseases of Zhejiang Province, Key Laboratory of Oral Biomedical Research of Zhejiang Province, Cancer Center of Zhejiang University},
            city={Hangzhou},
            postcode={310016}, 
            country={China}}  
            
\affiliation[five]{organization={Division of Periodontology and Implant Dentistry, Faculty of Dentistry, The University of Hong Kong},
            city={Hong Kong SAR},
            postcode={999077}, 
            country={China}}  

\affiliation[six]{organization={School of Computer Science and Engineering, Southeast University},
            city={Nanjing},
            postcode={211102}, 
            country={China}}  

\affiliation[seven]{organization={Department of Oral Implantology, School and Hospital of Stomatology, Shanghai Engineering Research Center of Tooth Restoration and Regeneration, Tongji University},
            city={Shanghai},
            postcode={200070}, 
            country={China}} 
            
\begin{abstract}

\textbf{Background:}Accurate tooth segmentation from cone beam computed tomography (CBCT) images is crucial for digital dentistry but remains challenging in cases of interdental adhesions, which cause severe anatomical shape distortion. 

\textbf{Methods:} 
 To address this, we propose a deep learning framework that integrates semantic and shape awareness for shape-preserving segmentation. Our method introduces a target-tooth-centroid prompted multi-label learning strategy to model semantic relationships between teeth, reducing shape ambiguity. Additionally, a tooth-shape-aware learning mechanism explicitly enforces morphological constraints to preserve boundary integrity. These components are unified via multi-task learning, jointly optimizing segmentation and shape preservation. 
 
 \textbf{Results:} 
Extensive evaluations on internal and external datasets demonstrate that our approach significantly outperforms existing methods.

\textbf{Conclusions:}
Our approach effectively mitigates shape distortions and providing anatomically faithful tooth boundaries.



\end{abstract}




\end{frontmatter}



\section*{Author contributions: CRediT}
\textbf{Zongrui Ji}: Writing – original draft, Visualization, Validation, Project
administration, Methodology, Investigation, Formal analysis, Data curation, Conceptualization. 
\textbf{Zhiming Cui}: Writing – review \& editing, Methodology.
\textbf{Na Li}: Writing – review \& editing, Data curation, Validation.
\textbf{Miaojing Shi}: Writing – review \& editing.
\textbf{Qianhan Zheng}: Data curation. 
\textbf{Ke Deng}: Writing – review \& editing.
\textbf{Jingyang Zhang}: Writing – review \& editing, Methodology.
\textbf{Xuepeng Chen}: Data curation, Validation, Writing – review \& editing. 
\textbf{Yi Dong}: Writing – review \& editing, Supervision, Funding acquisition. 
\textbf{Lei Ma}: Writing – original draft, Writing – review \& editing, Methodology, Conceptualization, Supervision, Funding acquisition.

\section*{Declaration of competing interest}
The authors declare that they have no known competing financial interests or personal relationships that could have appeared to
influence the work reported in this paper.

\section*{Funding sources}
This work was supported by National Natural Science Foundation of China (grant numbers 62473287, 62303320).

\section*{Data availability statement }
The data supporting the findings of this study are not publicly available at this time.
\section*{Permission to reproduce material from other sources}
All materials cited or reproduced from other sources in this study are in compliance with the principle of fair use, and no additional written permission from the copyright holders is required. All cited literatures and materials have been formally cited in the manuscript, with their original sources fully indicated.

\section{Introduction}
\label{introduction}

In digital dentistry, tooth segmentation from cone-beam computed tomography (CBCT) images is a critical process that produces the 3D models with faithful shape representation required for various procedures such as dental implantation and orthodontic realignment~\cite{R5}.
In clinical practice, dentists usually need to manually segment each tooth in the CBCT image slice by slice. This process is not only time-consuming and laborious, but also highly dependent on the experience of the operator \cite{park2025comparisons}. Therefore, it is of great importance to investigate an automated and accurate segmentation method for dental CBCT images. However, this task remains challenging due to complex geometric variations among teeth and, more critically, the prevalent adhesive boundaries between adjacent teeth which often lead to severe shape distortions in automated segmentation results. To cope with these challenges, researchers have proposed a variety of segmentation methods, which are mainly categorized into two types: traditional knowledge-based methods and deep learning-based methods.

Traditional knowledge-based methods leverage image processing techniques, mathematical modeling, and domain expertise to extract tooth structures by analyzing gray-scale features in CBCT images. 
These approaches primarily include threshold-based, edge-based, region-based, and contour-based segmentation methods~\cite{R5,R3,R4,R6,R7}.
Threshold-based methods segment images by comparing pixel gray values with predefined thresholds. For example, Marin et al. used an adaptive threshold model to filter images and identify tooth edges~\cite{y3}. Edge-based methods segment objects by detecting changes in grayscale, color, and texture. Pavaloiu et al. utilized the Canny operator combined with dental anatomical knowledge for edge detection, reducing computational burden and improving accuracy~\cite{y4,y5}. Region-based methods achieve segmentation by identifying homogeneous regions, focusing on contextual information to address the issue of blurred tooth boundaries. Contour-based methods primarily outline tooth boundaries or contours to segment complex morphological teeth from CBCT images ~\cite{y6,y7,y8}. However, these methods require extensive mathematical computations and must address issues related to metal artifacts and the separation of upper and lower teeth due to similar densities ~\cite{y9}. Despite their contributions, traditional methods are often hindered by poor accuracy, low reliability, and a dependence on manual operations, which limit their applications in clinical practice.

Recently, numerous studies have leveraged deep learning techniques to enhance the accuracy and efficiency of tooth segmentation. Deep learning-based methods achieve tooth delineation by hierarchically learning and integrating local-to-global features using semantic or instance segmentation approaches~\cite{R8}. The semantic segmentation model for teeth is primarily based on convolutional neural networks (CNNs). Ma et al combined a lightweight CNN with a classical level set method (referred to as the distance-regularized geodesic active contour model)~\cite{y10}. Wang et al. introduced an innovative multi-scale dense (MS-D) CNN, which merges scales within each layer and establishes dense connections between all feature maps ~\cite{y11}.
The aforementioned semantic segmentation methods lack the ability to perform more detailed analysis and operations on individual teeth, so many studies have introduced instance segmentation models for segmenting individual teeth. Instance segmentation methods first detect or locate the positions of different teeth and then independently outline each tooth.
For example, Cui et al. introduced a mixed-scale dense CNN that merges scales within each layer and establishes dense connections between all feature maps \cite{R9}. Instance segmentation models usually first detect or localize the position of different teeth, followed by a detailed understanding of the spatial boundaries and types of each tooth. Cui et al. developed a deep learning-based artificial intelligence system that is clinically stable and accurate for fully automated tooth and root segmentation from dental CBCT images \cite{R1}. Lv et al. designed a network that stably differentiates each tooth and captures geometric information to improve the accuracy of root segmentation \cite{R3}.

Existing deep learning methods for tooth segmentation still encounter significant limitations. Primarily, these approaches often disregard the inherent anatomical relationships between teeth, particularly neighboring ones, which are vital for achieving accurate segmentation. Secondly, the close physical proximity of adjacent teeth, coupled with their similar radiographic densities under natural occlusion, frequently causes mutual interference during the segmentation process. This interference hinders the accurate delineation of individual tooth boundaries and compromises the overall shape integrity of the segmentation results \cite{R17}.
Overcoming these obstacles necessitates novel segmentation strategies capable of effectively resolving spatial ambiguities and density-related confusions. Such advancements are crucial for achieving tooth segmentation outcomes that are not only accurate but also consistently preserve the faithful anatomical shapes of individual teeth.

In this study, to address these challenges, we develop a deep learning-based framework with semantic and shape awareness to achieve accurate tooth segmentation and prevent interdental adhesions. First, we propose a target-tooth-centroid prompted multi-label learning strategy to establish semantic awareness, explicitly modeling anatomical relationships between target and adjacent teeth to resolve boundary ambiguities and prevent interdental adhesions. Additionally, we incorporate a shape-aware learning mechanism that captures the morphological details of both the target tooth and its neighboring teeth, enabling accurate delineation of tooth boundaries across varying shapes. These dual perceptual components are unified within a multi-task architecture that jointly optimizes shape priors, mask segmentation, and boundary delineation via shared feature representation.
Collectively, the framework addresses spatial and structural complexities in CBCT data, delivering accurate segmentation with clinically viable shape preservation.

\begin{figure*}[t]
    \centering
    \includegraphics[width=0.95\textwidth]{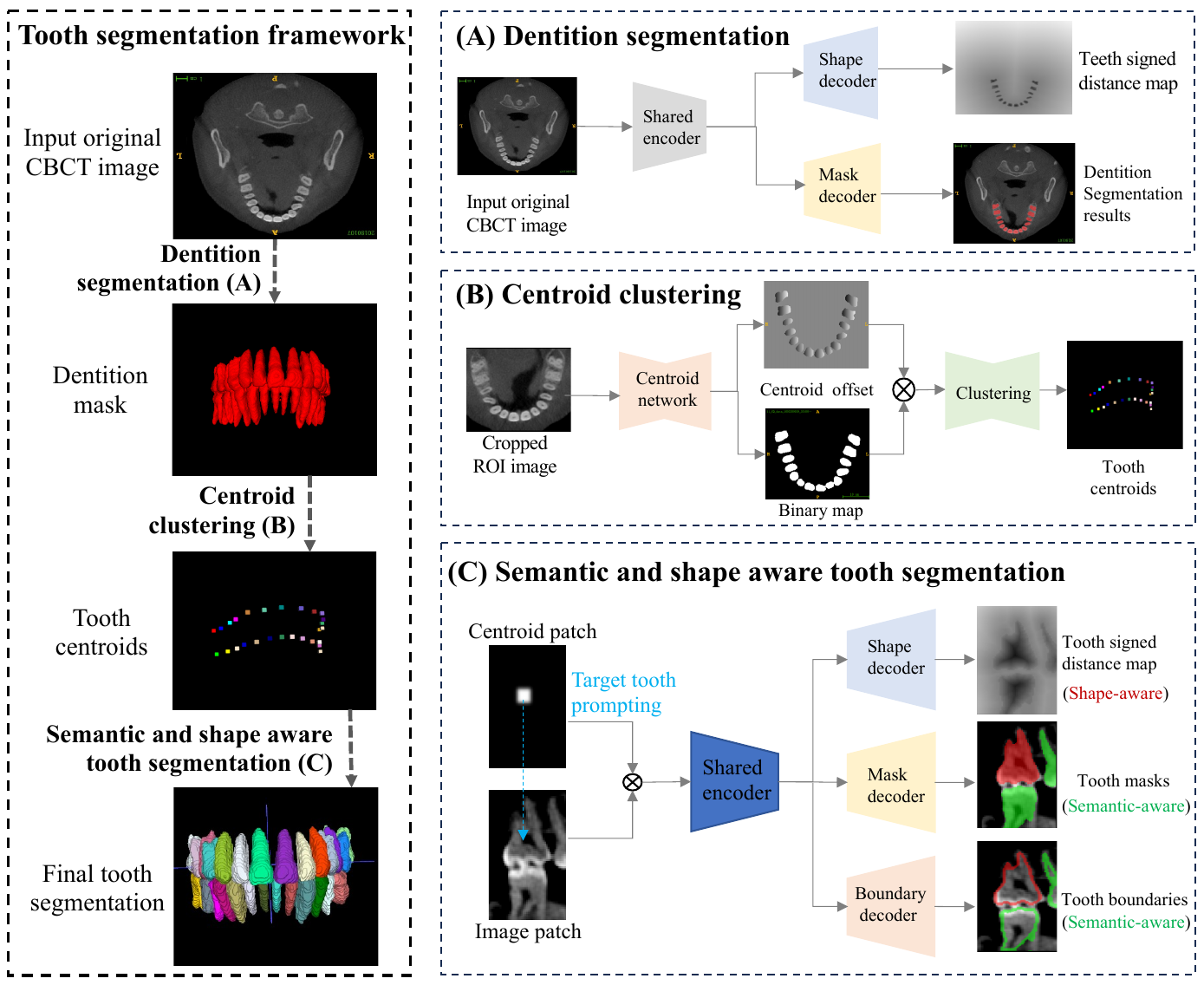}
    \caption{The architecture of the proposed tooth segmentation network. (A) The tooth region-of-interest (ROI) detection module. (B) The fast clustering based tooth centroid localization module. (C) the multi-task multi-label shape-aware segmentation module.}
    \label{F1}
\end{figure*}

\section{Materials and Methods}

\subsection{Network Architecture}

The architecture of the proposed framework is shown in Figure~\ref{F1}. 
Our framework consists of three progressive modules, i.e., a dentition segmentation module for tooth region-of-interest (ROI) detection, a centroid clustering module for individual tooth localization, and a semantic-aware and shape-aware tooth instance segmentation module.

\textbf{Dentition segmentation.} In the dentition segmentation module, we employ a U-Net-based dual-task segmentation network architecture, which achieves high-precision localization and boundary description of dentition through cooperative learning strategies. The network consists of a dentition mask segmentation sub-network and a dentition shape-aware sub-network to achieve multi-task joint optimization by sharing encoder features.
The mask segmentation sub-network extracts multi-scale features by the residual learning mechanism and the 3D convolution module of U-Net, while jump connections are utilized to retain the spatial information of different layers, which can effectively alleviate the problem of gradient disappearance.
The shape-aware sub-network utilizes a regression header to capture the shape features of the dentition by predicting its level set description. 
The level set description of dentition is defined using a signed distance map (SDM), which is expressed as the distance of a pixel to the object boundary. It assigns a positive or negative sign to the pixel depending on whether it lies inside or outside the object. This is calculated as follows:
\begin{equation}
T_r(y_i^a) = \begin{cases}
-\inf \min\limits_{y_i^b \in \partial S} \|y_i^a - y_i^b\|_2, & y_i^a \in S_{\text{in}} \\
0, & y_i^a \in \partial S \\
+\inf \min\limits_{y_i^b \in \partial S} \|y_i^a - y_i^b\|_2, & y_i^a \in S_{\text{out}} \\
\end{cases},
\end{equation}
where $y_i^a$ and $y_i^b$ denote different pixels in the segmentation truth $y_i$ and dentition boundary, respectively, and $\|y_i^a - y_i^b\|_2$ is the Euclidean distance between them. $N$ is the number of pixels. $S_\text{in}$ and $S_\text{out}$ are the inside and outside of the dentition, respectively, and $\partial S$ is the boundary of the dentition, which is regarded as the zero level set.

We define a shape prediction loss function that combines the boundary representation of the regression head and the topological embedding of the encoder as follows:
\begin{equation}
L_\text{shape} = \frac{1}{N} \sum_{i=1}^N L_\text{mse}(r_i, T_r(y_i)),
\end{equation}
where $r_i$ is the predicted level set obtained from the regression head using the image $x_i$ as input. $T_r(y_i)$ denotes the signed distance map computed over the segmentation ground truth $y_i$ and used as the underlying true level set function \cite{R18}. 

For the dentition mask segmentation sub-network, the segment loss function is used \cite{liu2021skullengine} combining the Dice and cross-entropy losses:
\begin{equation}
L_{\text{seg}} = \mu_1 L_{\text{seg\_dice}} + \mu_2 L_{\text{seg\_cross}}
\label{eq:seg_loss}
\end{equation}
where $L_{\text{seg\_dice}}$ and $L_{\text{seg\_cross}}$ represent the Dice loss and the cross-entropy loss, respectively. $\mu_1, \mu_2$ are weight parameters (typically $\mu_1 + \mu_2 = 1$). This combined loss balances robustness to imbalance and training stability.

The Dice loss is defined to measure the overlap between the predicted segmentation and the ground truth, expressed as:
\begin{equation}
L_{\text{seg\_dice}} = 1 - \frac{2|A \cap B|}{|A| + |B|}
\label{eq:dice_loss}
\end{equation}
where $A$ denotes the predicted segmentation region, $B$ represents the ground-truth label region. $|A \cap B|$ is the number of overlapping pixels, and $|A|$, $|B|$ are the pixel counts of the predicted and ground-truth regions, respectively. This loss is robust to class imbalance.

The cross-entropy loss measures the dissimilarity between the predicted probability distribution and the ground truth:
\begin{equation}
L_{\text{seg\_cross}} = -\frac{1}{M} \sum_{j=1}^{M} \left[ q_j \log(p_j) + (1 - q_j) \log(1 - p_j) \right]
\label{eq:cross_entropy}
\end{equation}
Here, $M$ is the total number of samples, $q_j \in \{0, 1\}$ is the ground-truth label of the $j$-th sample, and $p_j$ is the predicted probability of the positive class for the $j$-th sample. Cross-entropy provides stable gradients during the initial training phase.

The overall loss function $\mathcal{L}_{dentition}$ of the dentition segmentation network is formulated as a weighted summation of the shape prediction loss and dice loss as follows:
\begin{equation}
L_{\text{dentition}} = L_{\text{seg}} +\lambda_0 L_{\text{shape}},
\end{equation}
where $L_{\text{seg}},  L_{\text{shape}}$ denote the loss function for the segmentation mask, and shape task respectively. $\lambda_0$ represents the weights of $L_{\text{shape}}$, respectively.

By this design, the SDM results output from the shape segmentation sub-network provide a geometric prior for mask segmentation, while the binary masks generated by the mask segmentation sub-network provide regional constraints for shape regression.
This can effectively predict the boundaries and topology of dentition, further improving the segmentation accuracy.

\textbf{Centroid Clustering.}
As illustrated in Figure.~\ref{F1}, the centroid clustering module adopts the architecture proposed in \cite{R9}. Specifically, we first extract the region of interest (ROI) from the original CBCT image using the mask obtained from the dentition segmentation module. This cropped ROI is then processed by a centroid prediction network to generate both 3D centroid offsets and binary segmentation maps of teeth. 
These outputs serve as inputs to the centroid clustering module for generating single-tooth centroids. Within this module, we construct a tooth centroid density map by computing voxel frequencies. Applying a fast search clustering method~\cite{R21}, we detect density peaks to predict tooth centroids $T_c$. Each foreground voxel is assigned an instance label determined by its minimum distance to the predicted centroids in $T_c$. These instance labels function as: (1) anchor points for localizing individual tooth regions in the original images during instance segmentation, and (2) prompt inputs to the segmentation network in the subsequent module, ensuring precise and robust tooth segmentation.

\textbf{Semantic and Shape Aware Tooth Segmentation.}
In the third module of individual tooth segmentation, given a target tooth to be segmented, we first crop its sub-region from both the original image and the corresponding binary centroid map using the tooth's centroid. These two components are then concatenated to form a two-channel input.
The binary centroid serves as a prompt to explicitly direct network attention to the
target tooth region, thereby enhancing the segmentation accuracy of the target tooth. This centroid prompting is essential because the input image patch may contain multiple teeth, requiring explicit guidance for the network to identify the specific tooth to segment.
To address interdental adhesions, we integrate this target-tooth-centroid prompting with a multi-label learning strategy that explicitly models semantic relationships between the target tooth and adjacent teeth, establishing semantic awareness.
By integrating the centroid prompting with the multi-label learning, the network is able to more effectively learn the semantic relationships between the focal tooth and its neighbors, thus improving the tooth instance segmentation accuracy.
During training, we assigned distinct labels to the target tooth and its adjacent teeth, while correlating the input centroid with the target tooth label by setting them to identical values. 

To further enhance the morphological accuracy of segmented teeth, we leverage the shape-aware sub-network from the first module (dentition segmentation) to better preserve shape information of both the target tooth and adjacent teeth. 
The shape sub-network follows the network architecture and computational logic of the first stage, but its label assignment strategy is improved: two types of independent labels are assigned to the target tooth and its neighboring teeth, and a Signed Distance Map (SDM) is generated by the multi-label supervisory mechanism, which achieves a fine-grained geometrical description of the target object's Level Set. The design enhances the accuracy of the network in locating the boundary of tooth instances by explicitly distinguishing the spatial semantic relationship between the target teeth and the neighboring regions.

Building upon this foundation, we design a comprehensive multi-task learning framework that integrates three specialized components: the shape-aware sub-network for geometric feature extraction, a dedicated tooth mask segmentation sub-network for instance-level identification, and a tooth boundary delineation sub-network for precise contour mapping. 
The three sub-networks share a common encoder backbone, which significantly improves parameter efficiency while ensuring feature consistency across tasks. This shared design allows simultaneous extraction of complementary geometric, semantic and boundary features from the same latent representation.
The unified encoder employs a standard convolutional architecture to extract multi-scale features, while each sub-network utilizes dedicated decoder pathways: mask decoder reconstructs instance segmentation maps, boundary decoder focuses on contour refinement, shape decoder preserves morphological details
The network utilizes ReLU activations for nonlinear transformations and Softmax final layer for multi-category probability prediction (target tooth, adjacent teeth, background), addressing the requirements of dental segmentation tasks.
To ensure balanced optimization during the training phase, we implement an adaptive weighted loss function that dynamically adjusts the contribution weights across different learning tasks and label categories. This loss mechanism is formulated as a weighted summation of individual task-specific losses, where each task (shape perception, mask segmentation, and boundary delineation) is assigned distinct weighting coefficients based on its relative importance in the overall optimization objective. 
The overall loss function $\mathcal{L}_{total}$ is formally defined as follows:
\begin{equation}
L_{\text{total}} = L_{\text{multi\_seg}} + \lambda_1 L_{\text{multi\_bdr}} + \lambda_2 L_{\text{shape}},
\end{equation}
where $L_{\text{multi\_seg}}, L_{\text{multi\_bdr}}, L_{\text{shape}}$ denote the loss function for the segmentation mask, boundary, and shape task respectively. $\lambda_1$ and $\lambda_2$ represent the weights of $L_{\text{multi\_bdr}}$ and $L_{\text{shape}}$, respectively. $L_{\text{shape}}$ is defined following the equation (2). Both $L_{\text{multi\_seg}}$ and $L_{\text{multi\_bdr}}$ use the dice loss function and multi-label cross-entropy loss function.


The multi-label Dice loss for \( C \) classes is defined as:
\begin{equation}
L_{\text{multi\_dice}} = 1 - \frac{2\sum_{c=1}^{C} \omega_c |A_c \cap B_c|}{\sum_{c=1}^{C} \omega_c (|A_c| + |B_c|)}
\end{equation}
where
    \( A_c \) and \( B_c \) denote the predicted and ground-truth regions for class \( c \), respectively.
    \( |A_c \cap B_c| \) is the number of overlapping pixels for class \( c \).
    \( |A_c| \) and \( |B_c| \) are the pixel counts of the predicted and ground-truth regions for class \( c \).
    \( \omega_c \) weighted version assigns class-specific weights. 

For independent binary classification across \( C \) classes, the multi-label cross-entropy loss is:

\begin{equation}
L_{\text{multi\_cross}} = -\frac{1}{M} \sum_{j=1}^{M} \sum_{c=1}^{C} \omega_c \left[ q_{j,c} \log(p_{j,c}) + (1 - q_{j,c}) \log(1 - p_{j,c}) \right],
\end{equation}
where
    \( M \) is the number of samples (or pixels).
    \( q_{j,c} \in \{0, 1\} \) is the ground-truth label for sample \( j \) and class \( c \).
    \( p_{j,c} \in [0, 1] \) is the predicted probability of sample \( j \) belonging to class \( c \).
    \( \omega_c \) incorporates class weights  to mitigate imbalance.

\section{Results}

\subsection{Data and Preprocessing} 

\begin{figure*}[t]
    \centering
     \includegraphics[width=0.8\textwidth]{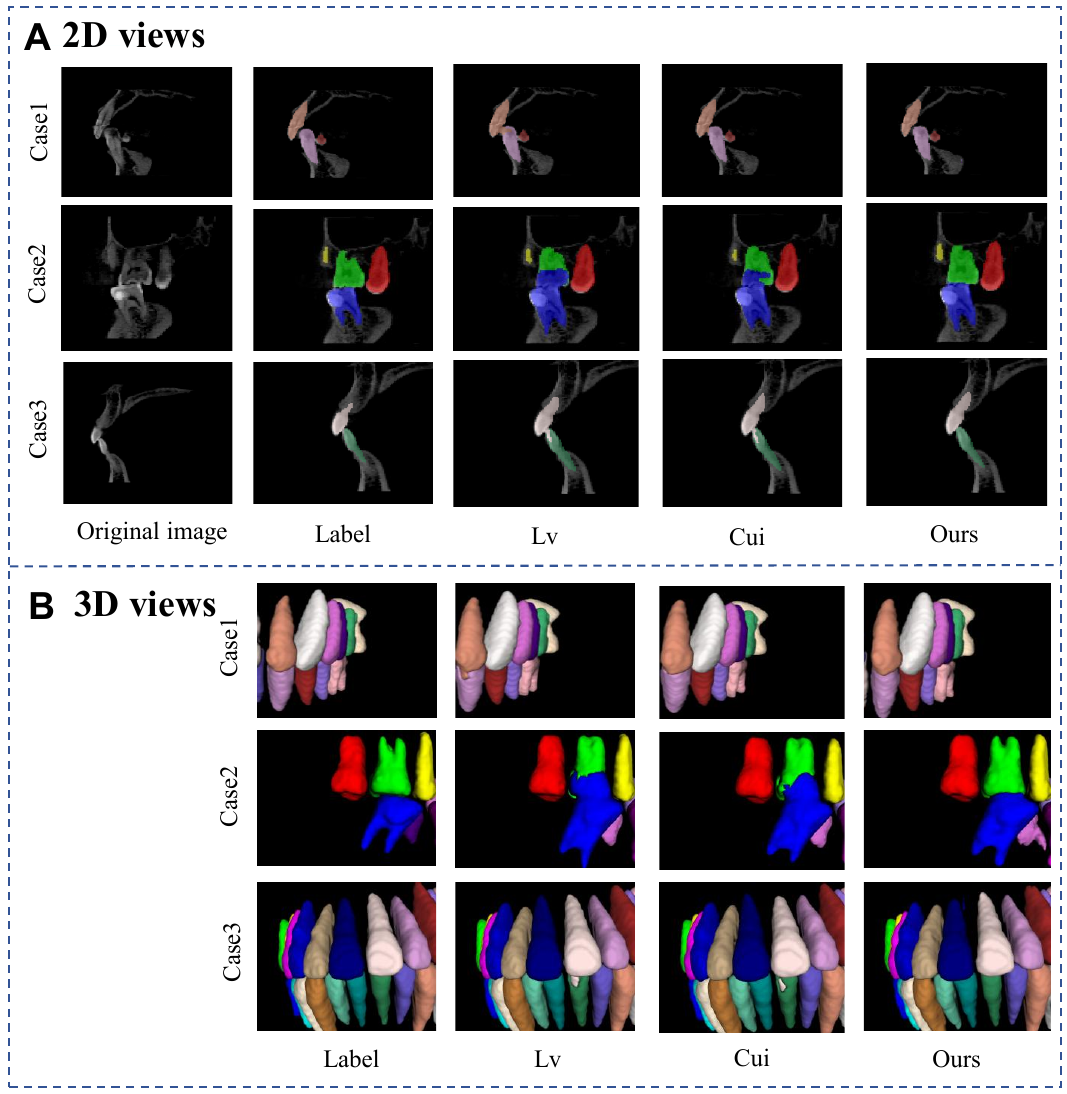}
    
    \caption{Visual comparison of tooth segmentation results using three different methods. Three typical examples are shown, each example displayed in both 2D (A) and 3D (B) views.}
    \label{F2}
\end{figure*}

To comprehensively develop and rigorously evaluate the proposed tooth segmentation framework, we systematically collected cone-beam computed tomography (CBCT) image data of patients who underwent treatment from multiple accredited medical centers. 
The dataset construction process involved two primary components as shown in Figure \ref{F2}. First, we assembled an internal dataset consisting of 827 high-quality CBCT images: 542 CBCT images from Henan
Stomatological Hospital, 225 CBCT images from Zhejiang University Stomatological Hospital and 60 CBCT images from public dataset ToothFairy \cite{da}.
To ensure robust model development and validation, this internal dataset was partitioned using a stratified random sampling approach. Specifically, 643 images were allocated to the training set, which was used to optimize the model's parameters, while 184 images were reserved for the testing set.
To further assess the clinical utility and generalization ability of the proposed model across diverse patient populations and imaging conditions, we incorporated an external dataset of 60 CBCT images from Tongji Stomatological Hospital, which had different scanning parameters, and patient characteristics from the medical institution in the internal dataset, providing a comprehensive evaluation of the model's adaptability in real-world clinical scenarios.
Prior to processing the 3D CBCT images with the deep learning network, we standardized the resolution to 0.4 × 0.4 × 0.4 mm³ to strike an optimal balance between computational efficiency and segmentation accuracy. 

\begin{figure*}[h]
    \centering
     \includegraphics[width=0.8\textwidth]{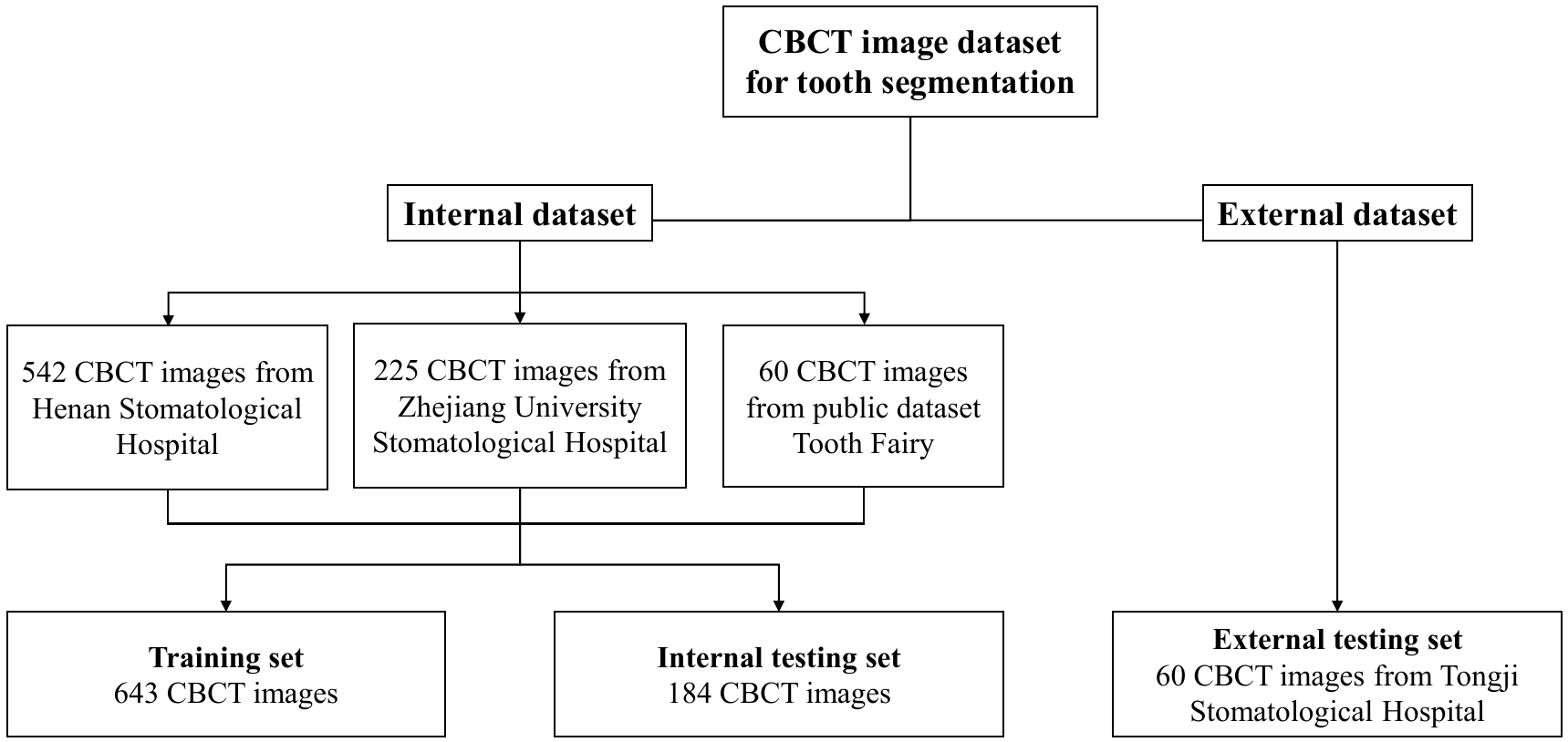}
    
    \caption{Composition of internal and external datasets for method development and evaluation.}
    \label{F4}
\end{figure*}

\begin{table*}
\caption{Quantitative results of different methods on the internal and external datasets. Dice, Jaccard, HD, and ASD represent Dice coefficient, Jaccard index, Hausdorff distance, and average surface distance, respectively.}\label{T0}
\small\centering
\begin{tabular}{|l|c|c|c|c|c|c|c|c|}
\hline
& \multicolumn{4}{c|}{\textbf{Internal Dataset}} & \multicolumn{4}{c|}{\textbf{External Dataset}} \\
\hline
Method & Dice & Jaccard & HD (mm) & ASD (mm) & Dice & Jaccard & HD (mm) & ASD (mm) \\
\hline
Lv \cite{R3} & 0.9229 & 0.8571 & 2.8448 & 1.4203 & 0.8812 & 0.7885 & 7.1099 & 1.5717 \\
Cui \cite{R1} & 0.9378 & 0.8835 & 1.2092 & 0.7835 & 0.8867 & 0.7969 & 5.7834 & 1.1033 \\
\textbf{Ours} & \textbf{0.9408} & \textbf{0.8885} & \textbf{1.1958} & \textbf{0.5142} & \textbf{0.8934} & \textbf{0.8083} & \textbf{2.3841} & \textbf{0.7719} \\
\hline
\end{tabular}
\end{table*}

\subsection{Experimental Setup}

We implemented the proposed framework using PyTorch and trained them on a NVIDIA GPU 3090 with 24 GB of memory. The training batch sizes for the first and second module networks were set to 1 and 4, respectively. For both networks, the initial learning rate was set to 0.001, with a decay factor of 0.8 applied every 2000 iterations, over a total of 30,000 iterations. For the third module network, the batch size was set to 8, with an initial learning rate of 0.001 and a decay factor of 0.8 every 200 iterations, totaling 6000 iterations. $\lambda_1$ was set to 0.1, and $\lambda_1, \lambda_2$ were both set to 0.4. All models were optimized using the Adam optimizer and trained to minimize the loss function.

In the model evaluation session, we first systematically compared our proposed framework with two current state-of-the-art networks in the literature \cite{R3} and \cite{R1}. To ensure the reliability and robustness of the evaluation results, this experiment was carried out on internal and external datasets, respectively, to validate the performance of the model under different data distributions and to measure its validity and generalization through multi-source data.
Second, to investigate the specific contribution of each network component in the proposed framework to the model performance, two sets of ablation experiments were designed and implemented in this study. The first set of experiments focuses on the frist dentition segmentation module, and verifies the effect of shape-aware learning on dentition segmentation accuracy by comparing two network configurations with and without shape-aware sub-network.
In the second set of experiments, four model variants were constructed for the third individual tooth segmentation module for comparative analysis: the baseline network (B-Net) did not introduce tooth centroid prompt, multi-label learning strategy, and shape-awareness, and retains only the basic segmentation function; C-net added only the tooth centroid prompt on top of the B-Net, which was used to guide the network to locate the target tooth region; CM-net further integrated the tooth centroid prompt and multi-label learning strategy to strengthen the ability to model the semantic relationship between the target tooth and the neighboring teeth; and our full network with shape awareness (CMS-Net) integrated the shape-aware learning on top of CM-net to achieve the accurate description of tooth geometry. By comparing these four network variants, the degree of contribution of each module to segmentation accuracy, boundary inscription and overall performance could be systematically evaluated.

\begin{figure*}[t]
    \centering
    \includegraphics[width=0.85\textwidth]{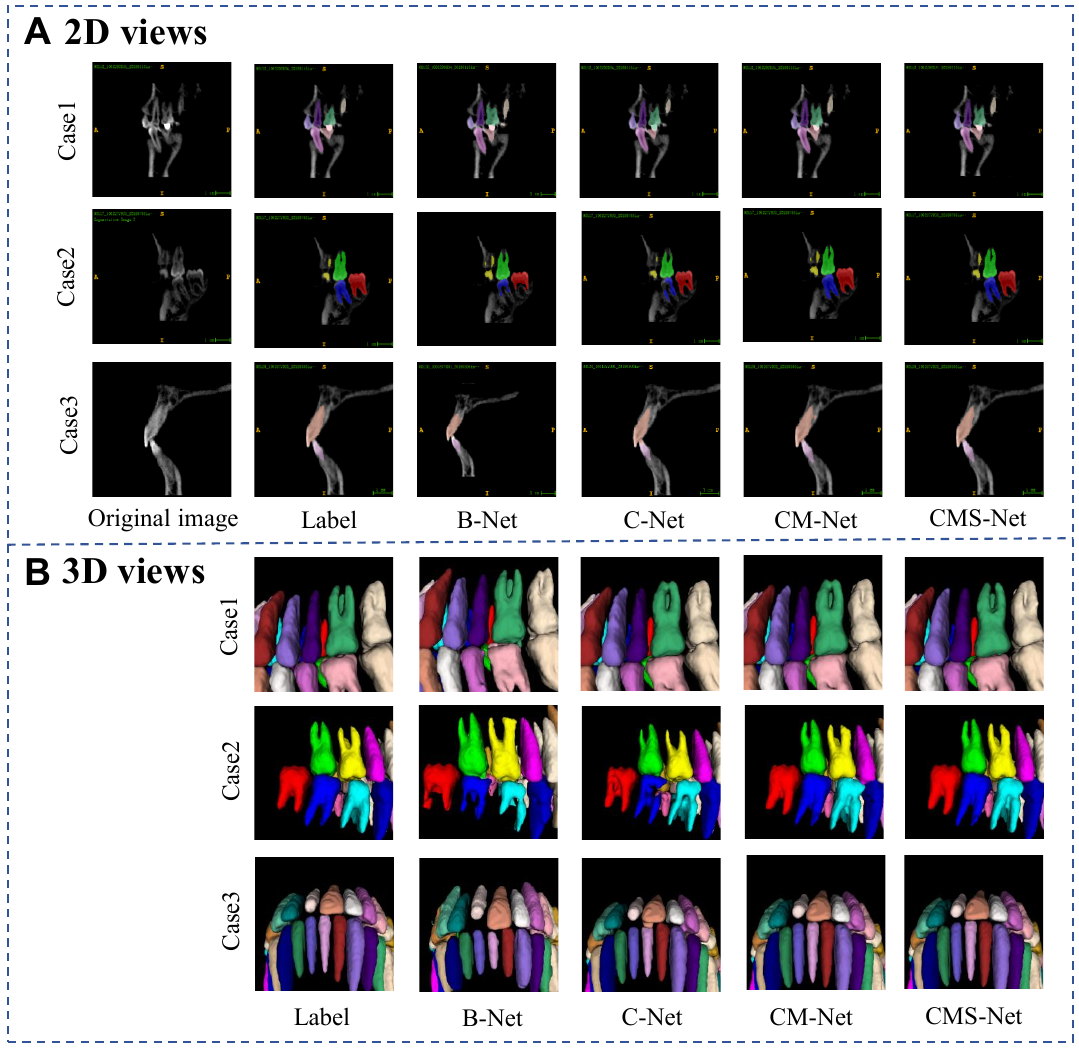}
    \caption{
    The qualitative comparison of tooth segmentation results in the ablation study includes the following networks: the baseline B-Net; C-Net, which incorporates the tooth centroid prompt into B-Net; CM-Net, formed by adding multi-label learning to C-Net; and the complete network (CMS-Net), integrated with shape-aware learning. Three representative examples are presented, each visualized in both 2D (A) and 3D (B) views.}
    \label{F3}
\end{figure*}

We conducted both quantitative and qualitative evaluations of the segmentation results for each network model.
For the quantitative evaluation, we used four metrics: the Dice coefficient, the Jaccard coefficient, the Hausdorff distance (HD), and the Average surface distance (ASD). Each metric evaluates the model’s segmentation performance from a different perspective. The Dice and Jaccard coefficients measured the overlap between the predicted and ground truth segmentations, with higher values indicating better performance. The HD and ASD assessed the accuracy of boundary prediction, with smaller values indicating lower deviations and greater precision.
Additionally, we performed a qualitative analysis by visually examining tooth adhesion and morphological integrity.
This multidimensional evaluation offered a comprehensive assessment of the model's segmentation performance and improvements.

\subsection{Results}
As shown in Table~\ref{T0}, our method outperforms the two comparison methods~\cite{R3,R1} across all metrics on the internal dataset. Specifically, our method achieves Dice and Jaccard coefficients of 0.9408 and 0.8885, respectively, surpassing~\cite{R3} (0.9229 and 0.8571) and~\cite{R1} (0.9378 and 0.8835). Additionally, our method demonstrates improved boundary alignment. The Hausdorff Distance (HD) for our method is 1.1958, which is lower than the values for~\cite{R3} (2.8448) and~\cite{R1} (1.2092). Similarly, our method achieves an average surface distance (ASD) of 0.5142, outperforming~\cite{R3} (1.4203) and~\cite{R1} (0.7835).
Figure~\ref{F2} presents three typical examples of tooth segmentation results achieved by the comparing methods. The visual comparisons highlight our method’s superior shape preservation and adhesion control, while the comparison methods struggle with complex shapes, resulting in more adhesion. These results effectively demonstrate the advantages of our method.

The evaluation results on the external dataset highlight strong generalization and robustness of our proposed method, suggesting its potential for clinical application. Specially, as shown in Table~\ref{T0}, our method outperforms the compared methods in both the Dice and Jaccard coefficients, achieving 0.8934 and 0.8083, respectively. This surpasses~\cite{R3} (Dice: 0.8812, Jaccard: 0.7885) and~\cite{R1} (Dice: 0.8867, Jaccard: 0.7969).
Additionally, our method excels in HD and ASD, achieving 2.3841 and 0.7719, respectively, while~\cite{R3} records 7.1099 for HD and 1.5717 for ASD, and~\cite{R1} shows 5.7834 for HD and 1.1033 for ASD.


\begin{table}[h]
\small\centering
\caption{Results of ablation studies on tooth centroid prompting, multi-label learning and shape-aware learning.\label{T2}}
\begin{tabular}{|l|c|c|c|c|}
\hline
Model    & Dice   & Jaccard & HD (mm) & ASD (mm) \\
\hline
B-Net    & 0.8737 & 0.7840  & 7.2717  & 2.2278  \\
C-Net    & 0.9164 & 0.8464  & 2.8420  & 1.5119  \\
CM-Net   & 0.9399 &0.8871  &1.4816  &1.0608
  \\
\textbf{CMS-Net}     & \textbf{0.9408} & \textbf{0.8885}  & \textbf{1.1958}  & \textbf{0.5142}  \\

\hline
\end{tabular}
\end{table}

\subsection{Ablation Studies}
To investigate the effect of shape-aware learning on the performance of the dentition segmentation in the first module, we conducted ablation experiments to comparatively analyze two networks with and without shape-aware sub-networks. The experimental results show that the network containing shape-aware sub-network outperforms the network without shape-aware sub-network in several key metrics.
The network with shape-aware sub-network has an average Dice coefficient of 0.9135 and an average Jaccard coefficient of 0.8416, while the network without shape-aware sub-network has an average Dice coefficient of 0.9097 and an average Jaccard coefficient of 0.8371. In terms of the metrics reflecting the boundary accuracy, the network with shape-aware sub-network has an average Hausdorff distance (HD) value of 2.5046 and an average surface distance (ASD) value of 2.1397; In contrast, the network without shape-aware sub-network had an average Hausdorff distance (HD) value of 5.5722 and an average surface distance (ASD) value of 2.2474.
The experimental results show that the shape-aware sub-network can improve accuracy of the dentition segmentation, which has a positive effect on the optimization of network performance.

We also conducted ablation studies to investigate the effectiveness of different components in the third individual tooth segmentation module.
The results of the ablation experiments in Table \ref {T2} and Figure \ref {F3} clearly present the layer-by-layer optimization effects of tooth centroid prompting, multi-label learning and shape-aware learning on tooth instance segmentation performance at the third module. The Dice coefficient of the basic network (B-Net) is 0.9097, the Jaccard coefficient is 0.8373, and the Hausdorff distance (HD) and average surface distance (ASD) are 2.7791mm and 1.075mm, respectively, which indicates that there is still room for optimization in the accurate segmentation of complex tooth structures.
After the introduction of the tooth centroid prompting to construct C-Net, the indexes are significantly improved: the Dice coefficient jumps to 0.9300, the Jaccard coefficient reaches 0.8704, and the HD and ASD are reduced to 2.1534mm and 0.9397mm, respectively, which is due to the fact that tooth centroid prompting can guide the network to accurately locate the target region and provide the key semantic anchors for the network, which greatly improves the segmentation efficiency and accuracy. Based on this, the multi-label learning is added to form CM-Net, and the segmentation performance is further improved, with the Dice coefficient increased to 0.9416, the Jaccard coefficient reached 0.8902, and the HD and ASD optimized to 1.5339 mm and 0.6518 mm. The multi-label learning strategy effectively enhances the segmentation capability of the network on tooth boundary discrimination by explicitly modeling the semantic associations between target teeth and their neighboring teeth. By explicitly modeling the semantic associations between the target and neighboring teeth, the multi-label network effectively enhances the network's ability to discriminate tooth boundaries and reduces the mis-segmentation problem caused by tooth adhesion.
Finally, our complete network (CMS-Net) with integrated shape-aware learning achieves optimal performance in all evaluation metrics: the Dice coefficient reaches 0.9445, the Jaccard coefficient is improved to 0.8953, and the HD and ASD are reduced to 1.2941 mm and 0.5927 mm, respectively. The design of the shape-aware sub-network based on signed distance field can accurately capture and describe the edge geometry of adjacent teeth, and further refine the tooth edges by constraining the network prediction with the shape a prior knowledge, which effectively avoids segmentation blurring due to the shape similarity, and improves the overall segmentation accuracy and structural integrity. 
Combined with the visualization results in Figure \ref {F3}, our full network CMS-Net shows sharper boundary delineation and more accurate semantic differentiation than other variants when dealing with complex tooth morphology, which strongly supports the effect of the synergistic effect of various modules on the quality of tooth segmentation.

\section{Discussion}

Accurate tooth segmentation from CBCT images is crucial for computer-aided dental procedures, such as implant planning. However, challenges persist due to uneven intensity distribution, unclear tooth boundaries, and substantial morphological variations. To overcome these obstacles, we have proposed a shape-preserving tooth instance segmentation framework designed to balance global topological coherence with local boundary clarity. Comprehensive evaluation have demonstrated its effectiveness in resolving tooth adhesion, preserving anatomical morphology, and achieving superior accuracy, clinical feasibility, and robustness.

A core innovation is the proposal and integration of target-tooth-centroid prompting with multi-label learning. This strategy provides explicit spatial cues to guide segmentation of specific teeth while establishing crucial semantic awareness. The predicted centroid acts as a stable anchor point for precise localization. Ensuring unique centroid-to-tooth correspondence allows the centroid to explicitly direct network attention to the target area. Multi-label learning further distinguishes the target tooth, adjacent teeth, and background. Importantly, the spatial precision from the centroid prompt enables multi-label learning to more effectively discern unique semantic features (e.g., shape, position) per tooth class, significantly enhancing differentiation of adjacent or morphologically similar teeth.

Secondly, we employ a shape-aware learning based on the signed distance field (SDF) to improve shape fidelity. This component is designed to capture the essential complementary aspects of shape: local boundaries and global topology. Utilizing regression heads on the network backbone, it leverages level-set methods~\cite{R15} – implemented via signed distance maps derived from ground truth segmentations – to effectively model the complex and variable shapes of teeth in CBCT data. This design maintains image resolution throughout the decoding process and captures intricate morphological details across multiple scales. The outcome is the preservation of morphological integrity within the segmentation results and a substantial reduction in tooth adhesion artifacts.

Further, we unify centroid prompting, multi-label learning, and shape-aware learning within a multi-task learning architecture. This integrated framework explicitly models target and neighboring teeth to resolve boundary ambiguities in adherent regions. It incorporates three complementary tasks sharing an encoder: shape-aware (SDM regression), mask segmentation, and boundary delineation. The synergy between these tasks is paramount. The centroid prompt and multi-label learning provide indispensable spatial guidance and semantic differentiation, while the SDF-derived shape prior imposes strong geometric constraints. Joint optimization via an adaptive weighted loss leverages shared features, enabling mutual reinforcement—shape information guides segmentation and boundaries, while boundary details enhance shape modeling. This approach is vital for precise tooth segmentation, evidenced by significantly reduced adhesion and improved boundary metrics in ablation studies.

Despite these promising results, several limitations warrant consideration for future work. First, while our study involved a considerable number of patients, the sample size may not fully capture the diversity of the target population, potentially introducing bias. Future efforts should aim to include more diverse cases and seek independent validation through large-scale clinical trials. Second, model generalization was assessed using only one external dataset. To thoroughly evaluate robustness across varied imaging protocols and patient demographics, testing on multiple independent external datasets is essential. Third, integrating the proposed system into a broader range of digital dental applications beyond orthodontics—such as dental implant planning or complete denture design—is crucial to demonstrate its real-world utility and validate performance in more complex clinical scenarios. Finally, to potentially enhance prediction accuracy further, research exploring the integration of patient-specific clinical information (e.g., gender, age, medical history) with dental imaging data within the deep learning training paradigm is needed.

\bibliographystyle{elsarticle-num}

\bibliography{tooth_segmentation}    

\end{document}